\documentclass[twocolumn,conference]{IEEEtran}
\IEEEoverridecommandlockouts
\usepackage[
pdfauthor={derajan},
pdftitle={How to do this},
pdfstartview=XYZ,
bookmarks=true,
colorlinks=true,
linkcolor=blue,
urlcolor=blue,
citecolor=blue,
pdftex,
bookmarks=true,
linktocpage=true,   
hyperindex=true
]{hyperref}
\usepackage{caption}
\usepackage{cite}
\usepackage{amsmath,amssymb,amsfonts}
\usepackage{algorithmic}
\usepackage{graphicx}
\usepackage{textcomp}
\usepackage{multirow}
\usepackage{xcolor}
\usepackage{siunitx}
\usepackage[bottom]{footmisc} 
\usepackage[numbers,sort]{natbib}
\begin{document}

\title{CAPformer: Compression-Aware Pre-trained Transformer for Low-Light Image Enhancement
}
\author{
    \IEEEauthorblockN{Wei Wang, Zhi Jin$^{*}$ }
    \IEEEauthorblockA{School of Intelligent Systems Engineering, Shenzhen Campus of Sun Yat-sen University,Shenzhen,Guangdong 518107, P.R.China}
    \IEEEauthorblockA{Guangdong Provincial Key Laboratory of Fire Science and Technology, Guangzhou 510006, China}
    \IEEEauthorblockA{wangw359@mail2.sysu.edu.cn, jinzh26@mail.sysu.edu.cn} 
}
\twocolumn[{
\renewcommand\twocolumn[1][]{#1}
\maketitle
\begin{center}
    \captionsetup{type=figure}
    \vspace{0.5cm}
    \includegraphics[width=1\textwidth]{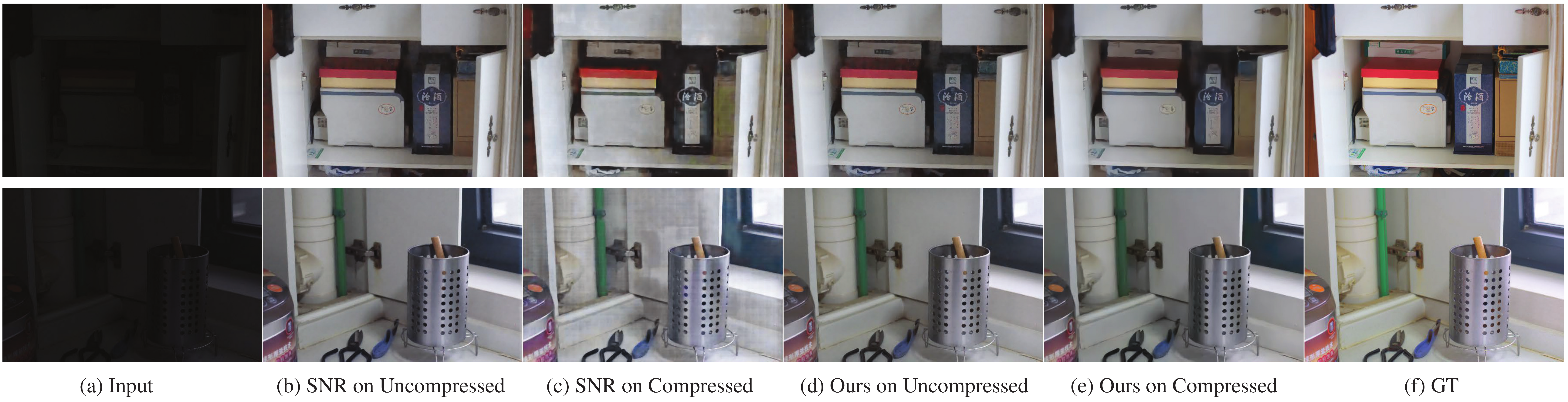}
    \captionof{figure}{The results on uncompressed and compressed low-light images of SOTA method SNR \cite{snr} and our CAPformer. Compression is set at JPEG QF80. SNR performs admirably on uncompressed images but yields unsatisfactory results when dealing with compressed ones. This trend is also observed in other SOTA methods, such as Retinexformer \cite{retinexformer}. In contrast, our method exhibits superior results on compressed images. Zoom in for a better view.}
    \vspace{0.5cm}
    \label{fig1}
\end{center}
}]

\begin{abstract}
Low-Light Image Enhancement (LLIE) has advanced with the surge in phone photography demand, yet many existing methods neglect compression, a crucial concern for resource-constrained phone photography. Most LLIE methods overlook this, hindering their effectiveness. In this study, we investigate the effects of JPEG compression on low-light images and reveal substantial information loss caused by JPEG due to widespread low pixel values in dark areas. Hence, we propose the Compression-Aware Pre-trained Transformer (CAPformer) network, employing a novel pre-training strategy to learn lossless information from uncompressed low-light images. Additionally, the proposed Brightness-Guided Self-Attention (BGSA) mechanism enhances rational information gathering. Experiments demonstrate the superiority of our approach in mitigating compression effects on LLIE, showcasing its potential for improving LLIE in resource-constrained scenarios.
\end{abstract}

\let\thefootnote\relax\footnotetext{* indicates the corresponding author. \\ Corresponding author : Zhi Jin (jinzh26@mail.sysu.edu.cn) 
}

\begin{figure*}[htbp]
\centering
\vspace{0.4cm}
    \includegraphics[width=1\textwidth]{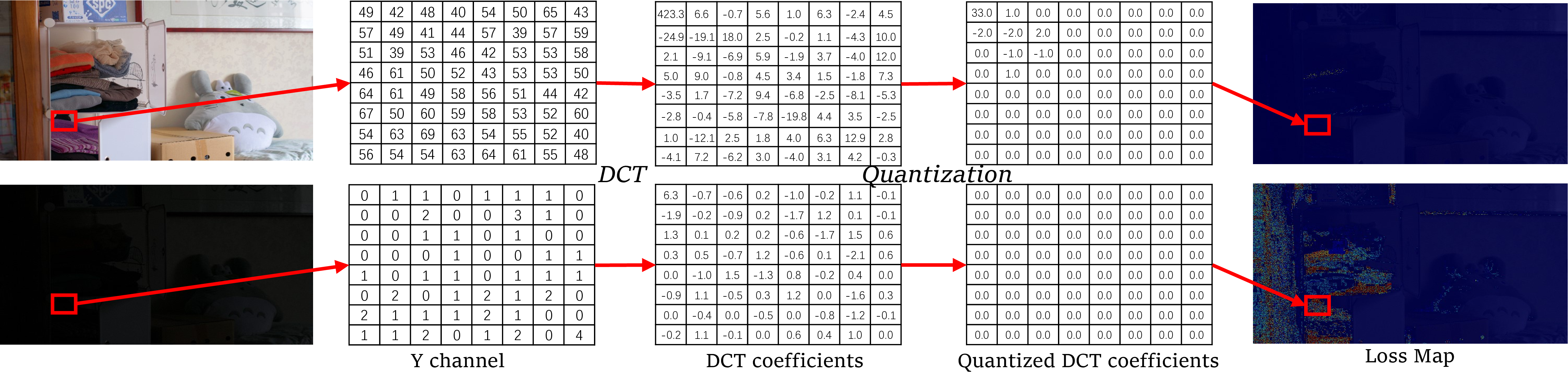}
\caption{The track of the JPEG process. The loss map on the right shows how severe the loss is. A higher density of colored areas indicates more significant information loss. Though compressed at a quite high QF of 80, the low-light image still suffers severe information loss concentrated in darker regions.}
\label{JPEG_Loss}
\end{figure*}


\section{Introduction}
Phone photography has emerged as a prevailing trend among consumers. However, due to inherent limitations in mobile hardware, the captured photographs often fails to meet user expectations, which has prompted the adoption of image enhancement algorithms to compensate for the hardware shortcomings. Low-Light Image Enhancement (LLIE) is an important technique to address the inherent challenges associated with nighttime photography. Methods like \cite{FourLLIE, kind+,retinexformer,snr,whjICME,brightenandcolorize} have achieved promising results. Besides advanced algorithms aimed at enhancing image quality, portability is another essential consideration in phone photography. Aiming to conserve storage space and facilitate photo transmission without unduly compromising image quality, compression is usually adopted after image acquisition. Hence, it is essential to expand LLIE research to tackle compressed issues. However, previous methods have predominantly omitted this critical aspect, and the handling of compressed images warrants substantial attention. As the most widely used image compression format, JPEG is chosen in this paper for further discussion and research. 

Many LLIE works have demonstrated impressive performance on commonly used datasets, which typically employ uncompressed images. However, the landscape takes a different shape when handling compressed images. As Fig.\ref{fig1}(c) shows, the previous SOTA method SNR \cite{snr} has failed to effectively address these JPEG compressed low-light images, even though the Quality Factor (QF) is set to a quite high one. Noise, artifacts, and pronounced smoothening deleteriously affect image quality, creating a great gap between the results on uncompressed ones (Fig.\ref{fig1}(b)). Similar situations have been observed in other SOTA methods like Retinexformer \cite{retinexformer}. This challenge comes from JPEG compression, where the loss incurred during quantization is irreversible because of the rounding operation, and it becomes particularly severe when dealing with low-light images due to the inherently small pixel values. Since dark regions are typically gathered to form a large area, this large area often suffers from severe loss, which may not be easily observed, but is challenging when the image is brightened. Existing methods have not considered this factor, resulting in their struggle for enhancement. 

To reconstruct these largely damaged areas, local information proves insufficient, requiring global information. Therefore, we design a U-shape network to extract features at different scales and a Transformer-based \cite{attentionisall} bottleneck to model long-range dependencies. Furthermore, we propose the Brightness-Guided Self-Attention (BGSA) mechanism, strategically guiding our model to disregard low-quality information originating from regions of extreme darkness. With the aforementioned designs, our network is adept at extracting information from compressed low-light images, enabling the reconstruction of severely damaged areas. Moreover, for such a task, an inclination is to initially restore the loss caused by compression before proceeding with LLIE. While such a cascaded plan may be effective, it often requires more parameters and training time, making it cumbersome. We aspire to achieve an enhancement from compressed low-light images to bright images directly. Inspired by the pre-training strategy of Transformer in low-level vision \cite{IPT, EPT}, we consider the recovery of compressed low-light images to uncompressed ones as the pre-training step. This enables the network to learn how to restore information in dark regions. As it has been pre-trained, our network has become more adept at reconstructing details in the dark of compressed images in LLIE. 

In this way, we propose the solution: Compression-Aware Pre-trained Transformer (CAPformer) for compressed low-light image enhancement. A pre-training strategy is applied, which is the first trial for LLIE. As shown in Fig.\ref{fig1}(e), our method has successfully achieved promising enhancement, especially compared to SNR \cite{snr}, eliminating the shortcomings present while significantly enhancing the fine details and textures. Further results are presented at EXPERIMENTS.

\begin{figure*}[htbp]
\centering
\vspace{0.4cm}
    \includegraphics[width=1\textwidth]{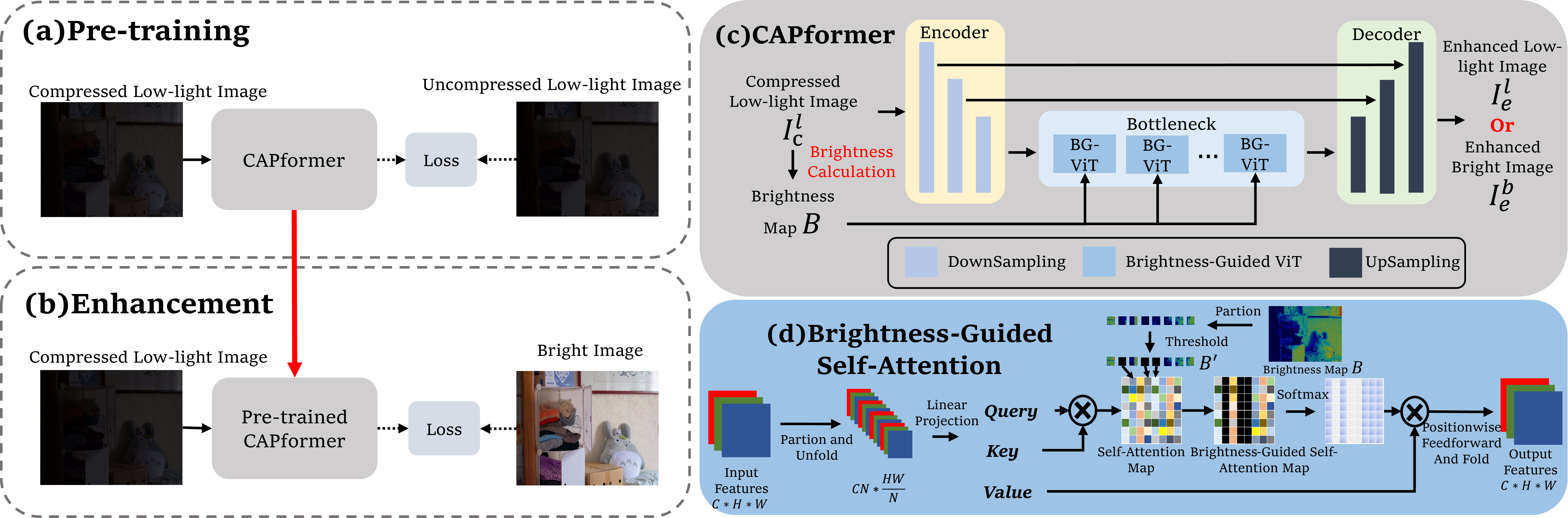}
\vspace{0.2cm}
\caption{Overview of the proposed method. CAPformer is pre-trained and then fine-tuned for enhancement. We employ a convolution layer with a stride of 2 for downsampling in the Encoder of CAPformer. The Decoder is symmetric to the Encoder, and the upsampling is implemented using the pixel shuffle layer. In BGSA, the black columns represent \num{-1e9}. After the $Softmax$ operation, these columns turn gray, representing nearly 0.}
\label{overview}
\end{figure*}

The main contributions are listed as follows:
\begin{itemize}
  \item [1)]
  To the best of our knowledge, we are the first to explore LLIE on compressed images with a learning method. After thorough validation, we confirm that JPEG compression indeed poses significant challenges for existing LLIE works. It deserves more attention and we provide our solution.
  \item [2)]
  We propose a novel pre-training method to learn lossless information from uncompressed low-light images before fine-tuning for LLIE. As far as we know, such a strategy has not been applied in LLIE yet.
  \item [3)]
  Due to the characteristics of compressed low-light images, we design a U-shape network with a Transformer-based bottleneck and propose a novel Brightness-Guided Self-Attention (BGSA) mechanism to model valid long-range dependencies.
  \item [4)]
  Experimental results demonstrate a substantial improvement compared to existing LLIEs on JPEG images and prove our method depresses the effect of compression successfully.
\end{itemize}

\section{Related works}
\subsection{Low-Light Image Enhancement}
Low-light image enhancement has gained significant attention in recent years, aiming to address visibility challenges and noise effects in nighttime scenes. With the rapid progress of deep learning, existing LLIE methods fall into two main categories: end-to-end frameworks and Retinex-based frameworks.

Retinex-based methods rely on the Retinex theory \cite{retinextheory}, decomposing images into reflectance and illumination components. Successive approaches, such as Retinex-Net \cite{LOLv1} and innovations like \cite{kind+} and Retinexformer \cite{retinexformer}, have integrated deep learning and Transformer architecture for long-range modeling. In contrast, end-to-end methods \cite{snr, mirnet, whjICME,FourLLIE,he2021srdrl,jin2019flexible,jin2017robust} learn mappings between low-light and normal-light images for enhancement. While early attempts like \cite{llnet} paved the way, recent methods fall into two categories: those leveraging deep learning's fitting capabilities and those incorporating physical concepts for interpretability.

Notably, existing learning-based approaches neglect compressed images, proving ineffective when confronted with them. While previous attempts \cite{tip18-related,eccv14-related,pcm18-related} have addressed JPEG compressed low-light images, relying on traditional algorithms, their performance lags behind deep learning methods that do not explicitly account for JPEG compression effects.

\subsection{JPEG Compression}
JPEG currently stands as the most prevalent format for storing and transmitting photos, striking a balance between storage efficiency and image quality preservation. The fundamental design of JPEG involves discarding redundant high-frequency components. As depicted in Fig.\ref{JPEG_Loss}, when transitioning to the frequency domain, JPEG utilizes quantization to eliminate high-frequency components. Among all the processes, quantization plays a predominant role in loss and may lead to noticeable artifacts.

Some works on JPEG artifact removal like \cite{qgac,fbcnn} incorporated JPEG priors to reduce quantization loss. However, besides high-frequency information, low-frequency information of dark areas also suffers from severe loss as shown in Fig.\ref{JPEG_Loss}. When subjected to the same scale of quantization and rounding operation, the loss is more severe compared to bright images as the loss map displays, posing a greater challenge for low-light image enhancement.

\subsection{Pre-training Transformer}
Pre-training Transformer \cite{attentionisall} has been proven effective and essential. IPT \cite{IPT} pioneered its application in low-level vision, achieving notable results through pre-training on ImageNet for denoising and super-resolution tasks. EDT \cite{EPT} further explores Transformer architecture principles during pre-training, introducing innovative paradigms for multi-task pre-training and achieving enhanced performance. However, these efforts primarily focused on low-cost synthetic degradation tasks and did not address complex degradations like low-light conditions.

Fortunately, our task involves two types of degradation, with one being JPEG compression, which can be easily simulated. This enables us to follow the paradigm of previous work and conduct pre-training for Transformer.

\begin{table*}[htbp]
    \centering
    \vspace{0.2cm}
    \caption{Experiments on LOL \cite{LOLv1}-JPEG datasets and LOLv2-Real \cite{LOLv2}-JPEG and LOLv2-Synthetic \cite{LOLv2}-JPEG datasets. ↑/↓ denotes that larger/smaller values lead to better quality. The bold denotes the best and the underline denotes the second best. }
    \resizebox{\textwidth}{0.14\textwidth}{
        \begin{tabular}{c|c|ccc|ccc|ccc} \hline
            \multirow{2}{*}{Methods}           & \multirow{2}{*}{Params(M)}           &
            \multicolumn{3}{c|}{LOLv1-JPEG}    & \multicolumn{3}{c|}{LOLv2-Real-JPEG} & \multicolumn{3}{c}{LOLv2-Synthetic-JPEG}                                                                                                                       \\
                                               &                                      & PSNR(dB)↑                                & SSIM↑              & LPIPS↓            & PSNR(dB)↑       & SSIM↑              & LPIPS↓ & PSNR(dB)↑ & SSIM↑ & LPIPS↓ \\ \hline
            MIRNet \cite{mirnet}               & 31.76
                                               & 21.3152                              & 0.7769                                   & 0.1772
                                               & 20.4468                              & 0.7677                                   & 0.1957
                                               & 22.4171                              & 0.8261                                   & 0.1378                                                                                                              \\
            KinD++ \cite{kind+}                & 8.27
                                               & 19.2118                              & 0.7749                                   & 0.1712
                                               & 16.9173                              & 0.7397                                   & 0.2358
                                               & 21.1032                              & 0.8054                                   & 0.1493                                                                                                              \\
            HWMNet \cite{HWMNet}               & 66.56
                                               & 20.7501                              & 0.7531                                   & 0.2279
                                               & 19.4828                              & 0.7449                                   & 0.3161
                                               & 22.0571                              & 0.8159                                   & 0.1365                                                                                                              \\
            SNR \cite{snr}                     & 39.13
                                               & 22.2014                              & 0.7759                                   & 0.1798
                                               & 20.6297                              & 0.7616                                   & 0.1656
                                               & 22.1376                              & 0.8246                                   & 0.1256                                                                                                              \\

            Restormer \cite{restormer}         & 26.13
                                               & 22.4093                              & 0.7739                                   & 0.1703
                                               & 21.0392                              & 0.7673                                   & 0.1864
                                               & 22.4908                              & 0.8227                                   & 0.1295                                                                                                              \\

            DiffLL \cite{diffLLIE}             & 22.08
                                               & 21.8140                              & 0.7612                                   & 0.1836
                                               & 18.3842                              & 0.6618                                   & 0.3282
                                               & 19.3190                              & 0.7027                                   & 0.2608                                                                                                              \\

            FourLLIE \cite{FourLLIE}           & 0.12
                                               & 20.6441                              & 0.7452                                   & 0.2186
                                               & 19.9017                              & 0.7491                                   & 0.2822
                                               & 21.7369                              & 0.8014                                   & 0.1730                                                                                                              \\

            Retinexformer \cite{retinexformer} & 1.61
                                               & 22.7566                              & 0.7791                                   & 0.1721
                                               & 21.0643                              & 0.7729                                   & 0.1871
                                               & 22.5854                              & 0.8255                                   & 0.1334                                                                                                              \\

            FBCNN \cite{fbcnn}→SNR             & 71.92+39.13
                                               & 21.7915                              & 0.7776                                   & 0.1642
                                               & 20.3031                              & 0.7745                                   & \textbf{0.1419}
                                               & 20.9890                              & 0.7898                                   & 0.1260                                                                                                              \\

            FBCNN \cite{fbcnn}→Restormer       & 71.92+26.13
                                               & 22.5208                              & 0.7891                                   & 0.1488
                                               & \underline{21.3024}                  & \underline{0.7939}                       & 0.1699
                                               & 23.0956                              & 0.8370                                   & 0.1258                                                                                                              \\

            FBCNN \cite{fbcnn}→FourLLIE        & 71.92+0.12
                                               & 21.0009                              & 0.7718                                   & 0.1760
                                               & 20.4363                              & 0.7642                                   & 0.2380
                                               & 21.9812                              & 0.8106                                   & 0.1726                                                                                                              \\

            FBCNN \cite{fbcnn}→Retinexformer   & 71.92+1.61
                                               & \underline{22.8734}                  & \underline{0.7931}                       & \underline{0.1472}
                                               & 20.8648                              & 0.7871                                   & 0.1814
                                               & \underline{23.1502}                  & \underline{0.8375}                       & \textbf{0.1176}                                                                                                     \\  \hline

            \textbf{CAPformer(Ours)}           & 8.77
                                               & \textbf{23.4993}                     & \textbf{0.8073}                          & \textbf{0.1455}
                                               & \textbf{21.6894}                     & \textbf{0.7965}                          & \underline{0.1640} & \textbf{23.2957 } & \textbf{0.8395} & \underline{0.1237}                                       \\ \hline
        \end{tabular} }
    \vspace{0.2cm}
    \label{table:1}
\end{table*}
\section{Method}
\subsection{Analysis}
To address the challenge of enhancing compressed low-light images, it is crucial to understand the reason for this challenge compared to working with uncompressed images. When applying JPEG compression to an extremely low-light image, substantial losses occur and are particularly hidden in darker regions, as illustrated in the loss map of Fig.\ref{JPEG_Loss}. Due to the small values, low-frequency components in the dark easily disappear during the quantization and rounding operations, making recovery challenging.
Therefore, the key point is to restore the information in dark areas. Dark regions are often gathered together to form large expanses of darkness, which indicates that the entire large region suffers from severe loss. Reconstructing such regions solely based on their information is challenging, necessitating help from more distant regions. Effectively modeling long-range dependencies becomes crucial, prompting the adoption of Transformer \cite{attentionisall} for this goal. As shown in Fig.\ref{overview}(c), we adopt a U-shape network to extract features from different scales, with the bottleneck consisting of several ViTs. 

Merely employing this network is not sufficient. Our proposed pre-training strategy and Brightness-Guided ViT can further enhance the performance.

\subsection{Brightness-Guided Self-Attention}
Relying on the original ViT \cite{ViT} proves inadequate for our task. In extremely dark regions of a JPEG compressed image, there is substantial loss of information and increased noise, and incorporating information from such areas may potentially degrade performance. Previous work \cite{snr} introduces a SNR map to guide attention and prevent excessive focus on low-quality information. However, the SNR map is unsuitable for our task, as SNR values in compressed low-light images are generally low, hindering effective discernment between high-quality and low-quality information.
Thus, as described in Fig.\ref{overview}(d), we propose our Brightness-Guided Self-Attention (BGSA) mechanism to avoid the propagation of low-quality information from those regions. We incorporate BGSA with standard ViT then we have BG-ViT.
We calculate the brightness map as follows:
\begin{equation}
    \mathcal{B} = \frac{gray(I_{c}^l)}{max(gray(I_{c}^l))},
\end{equation} where $gray(*)$ represents the calculation of gray image from original input.

In BGSA, attention is directed away from positions with low values in $\mathcal{B}$. This is accomplished by thresholding values in $\mathcal{B}$; those below the threshold are set to 0, and those above are set to 1, yielding $\mathcal{B^{\prime}}$. The BGSA is formulated as follows:
\begin{equation}
    \text { Attention }\left(\mathbf{Q}, \mathbf{K}, \mathbf{V}\right)=\operatorname{Softmax}\left(\mathbf{Q} \mathbf{K}^T+\left(1-\mathcal{B^{\prime}}\right) \sigma\right) \mathbf{V},
\end{equation}
 where Q, K, and V represent Query, Key, and Value in self-attention, respectively, and $\sigma$ is set to \num{-1e9} to drive the values in corresponding positions close to zero after the $Softmax$ function. 
 The effectiveness of the BGSA is proved in the ablation study.


\subsection{Pre-training and Enhancement}
The effectiveness of pre-training Transformer has been demonstrated and applied to several low-level vision tasks, excluding LLIE due to the complex degradation that is challenging to synthesize on large-scale datasets. As clarified by EDT \cite{EPT}, pre-training Transformer on related tasks may provide substantial assistance. Restoring information for the compressed low-light image is essentially a sub-task within the overall enhancement process, highly pertinent to the LLIE task. Therefore, setting this task as the pre-training of our Transformer-based network is a reasonable approach. The entire process is depicted in Fig.\ref{overview}(a) and Fig.\ref{overview}(b).

In the Pre-training stage, the proposed network is trained with the uncompressed low-light images as ground truth. We adopt Charbonnier loss \cite{charbonnier} as the loss function which can be written as
\begin{equation}
    L_p=\sqrt{\left\|f(I_{c}^l)-I_{u}^l\right\|_2+\epsilon^2},
\end{equation} where $I_{c}^l$ is the compressed low-light image and $f(*)$ represents the output of our model. $I_{u}^l$ means uncompressed low-light image. $\epsilon$ is set as \num{1e-3}.

Then in the Enhancement stage, uncompressed bright images are utilized as ground truth for fine-tuning. The Pre-training stage provides valuable insights, enabling the model to effectively enhance images by preserving texture and detail while illuminating them. The loss function is the same as the Pre-training stage:  
\begin{equation}
    L_e=\sqrt{\left\|f(I_{c}^l)-I_{u}^b\right\|_2+\epsilon^2},
\end{equation} which differ with the above one at $I_{u}^b$, the uncompressed bright image.
The effectiveness of this pre-training strategy is proved in the ablation study.

\begin{figure*}
    \centering
    \vspace{0.2cm}
    \includegraphics[width=1\textwidth]{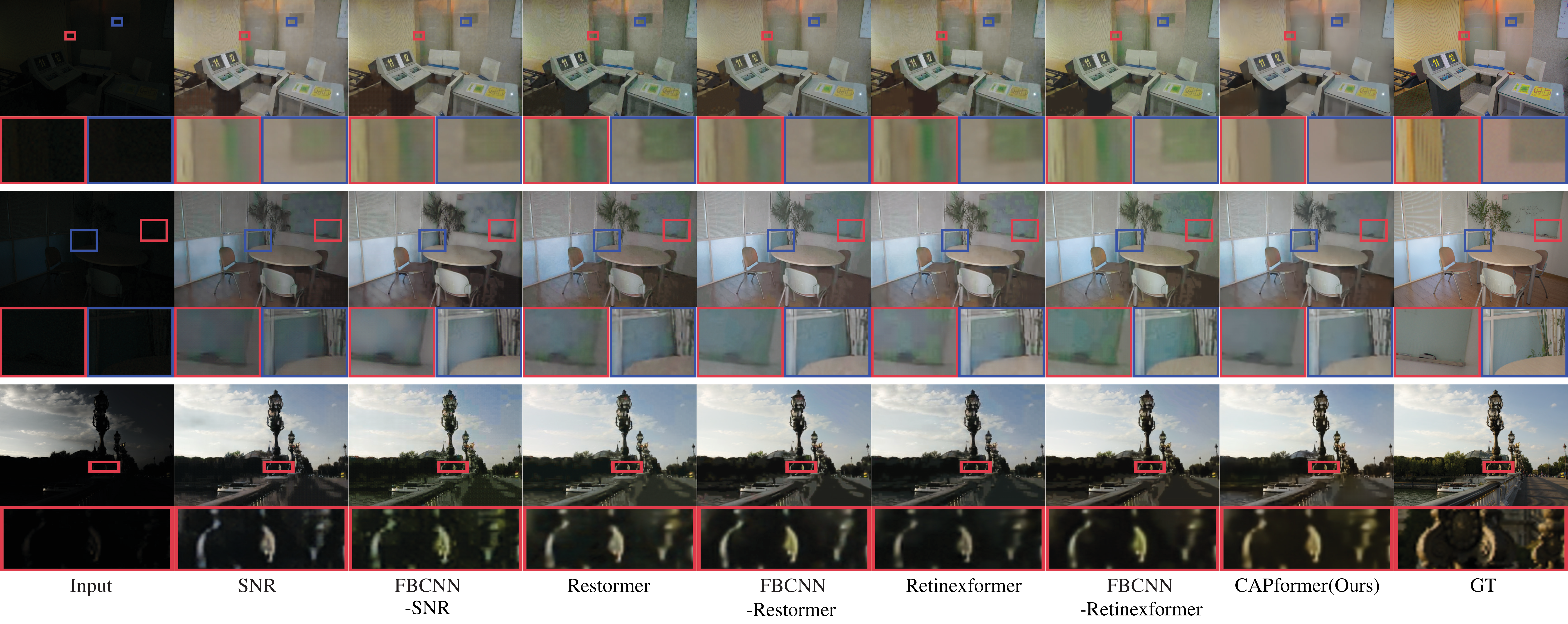}
    \caption{Visual comparison on LOLv1-JPEG, LOLv2-Real-JPEG, and LOLv2-Synthetic-JPEG (top to bottom). All inputs are compressed at QF 80. Our method yields fewer artifacts and better color consistency than others.}
    \vspace{0.3cm}
    \label{vision}
\end{figure*}

\section{Experiments}
\label{experi}
\subsection{Experimental Settings}
As recommended in \cite{qfrec}, setting the JPEG QF to 80 is considered representative of most daily situations due to its balance between storage efficiency and good visibility. Consequently, all our experiments were conducted on JPEG low-light images with QF 80. LOLv1 \cite{LOLv1}, LOLv2-Real \cite{LOLv2}, and LOLv2-Synthetic \cite{LOLv2} are selected as the fundamental LLIE datasets. The U-shaped network employs 3 downsampling layers, and the bottleneck contains 4 BG-ViTs. The threshold in the proposed BGSA is set to \num{1e-3}. The total number of parameters is $8.77M$.

\subsection{Quantitative Evaluation}
We conduct a comparative analysis of our method against LLIE works and restoration methods. For better comparison, we also employ cascaded approaches that integrate the leading method in JPEG artifact removal, namely FBCNN \cite{fbcnn}, to restore compressed low-light images before applying enhancement. All these approaches were retrained on JPEG-compressed datasets to ensure fairness. PSNR, SSIM \cite{ssim}, and Learned Perceptual Image Patch Similarity (LPIPS) \cite{lpips} are selected for performance assessment.

The results are listed in Table.\ref{table:1}. 
In terms of PSNR and SSIM \cite{ssim}, our proposed method achieves the best results across all three datasets. Regarding LPIPS \cite{lpips}, we achieve the best performance in LOLv1-JPEG and the second-best in LOLv2-Real-JPEG and LOLv2-Synthetic-JPEG. The pre-training process and the devised BGSA contribute to the ability to restore lost information and gather pertinent details for effective reconstruction. These results substantiates the effectiveness of our approach compared to alternative methods.

While the cascaded approaches generally provide improvement to the original methods, they still lag behind our approach after employing a substantial number of parameters. Even the most advanced JPEG restoration method cannot achieve complete recovery without loss, and the persistent unrecovered losses pose ongoing challenges for the original method. This inherent limitation of the cascaded approach underscores the efficacy and simplicity of our proposed method.

\subsection{Qualitative Evaluation}
We present visual samples in Fig.\ref{vision} for comparing our method with SNR \cite{snr}, Restormer \cite{restormer}, Retinexformer \cite{retinexformer} and their corresponding cascaded approaches that achieve the best performance (in terms of PSNR) on LOLv1-JPEG, LOLv2-Real-JPEG, and LOLv2-Synthetic-JPEG datasets. We select some patches in the dark to zoom in for better observation. Affected by compression, the results of other methods exhibit severe artifacts and noises in the top two rows, along with significant color shifts in the third row. While the cascaded plans partially alleviate this situation, they still trail behind our method. Our result demonstrates superior visual quality, with significantly fewer artifacts and noise, and better color consistency, affirming the effectiveness of our method in mitigating the effects of compression.

\subsection{Ablation Study}
We conduct five ablation experiments by individually removing or replacing different components. The evaluations are performed on LOLv1\cite{LOLv1}-JPEG at QF 80.

The baseline is established by omitting the pre-training strategy $\mathcal{P}$ and the BGSA $\mathcal{B}$ from BG-ViT. Additionally, we substitute the brightness map in BGSA $\mathcal{B}$ with the SNR map $\mathcal{S}$ to further assess our designed BGSA.

As indicated in Table.\ref{abst}, the exclusion of $\mathcal{B}$ and $\mathcal{P}$ detrimentally affects performance, affirming the effectiveness of both designs. The incorporation of the SNR map $\mathcal{S}$ has a negative impact on performance, attributed to its inability to discern effectively between high-quality and low-quality information in compressed low-light images.
\section{Conclusion}
In this paper, we identify the challenges faced by existing LLIE methods in handling compressed images and propose a comprehensive solution. Leveraging the characteristics of compressed low-light images, we design a U-shaped network with a Transformer-based bottleneck to capture reliable global attention. Moreover, the BGSA is introduced to further enhance global attention. Furthermore, we introduce a pre-training strategy that involves training our model with uncompressed low-light images before fine-tuning it for LLIE. Experimental results on compressed LLIE datasets demonstrate that our method surpasses SOTA approaches and effectively mitigates the impact of compression.
\begin{table}[]
\centering
\caption{Ablation Studies on LOLv1\cite{LOLv1}-JPEG. The bold denotes our complete CAPformer performing the best.}
\begin{tabular}{cccc|ccc} \hline
\multicolumn{1}{l|}{\multirow{2}{*}{Baseline}} & \multicolumn{1}{l|}{\multirow{2}{*}{$\mathcal{P}$}} & \multicolumn{1}{l|}{\multirow{2}{*}{$\mathcal{B}$}} & \multicolumn{1}{c|}{\multirow{2}{*}{$\mathcal{S}$}} & \multicolumn{3}{c}{LOLv1-JPEG} \\
\multicolumn{1}{l|}{}                          & \multicolumn{1}{l|}{}                   & \multicolumn{1}{l|}{}                   & \multicolumn{1}{c|}{}                   & PSNR(dB)↑      & SSIM↑     & LPIPS↓   \\ \hline
$\checkmark$  &      &               &               & 22.7545   & 0.7913   & 0.1647  \\
$\checkmark$  &      &               & $\checkmark$  & 22.3950   & 0.7920   & 0.1563  \\
$\checkmark$  &      & $\checkmark$  &               & 22.7691   & 0.7956   & 0.1580   \\
$\checkmark$  & $\checkmark$ &       &               & 23.0646   & 0.7964   & 0.1610  \\
$\checkmark$  & $\checkmark$ &       & $\checkmark$  & 22.7532   & 0.7974   & 0.1643  \\ \hline
$\checkmark$  & $\checkmark$ & $\checkmark$ &        
& \textbf{23.4993} & \textbf{0.8073}   & \textbf{0.1455}  \\ \hline
\end{tabular}
\vspace{0.3cm} 
\label{abst}
\end{table}

\section*{Acknowledgment}

This work was supported by the National Natural Science Foundation of China under Grant No. 62071500; Supported by Shenzhen Science and Technology Program under Grant No. JCYJ20230807111107015.

\bibliographystyle{ieeetr}
\bibliography{bib}

\begin{thebibliography}{10}

\bibitem{snr}
X.~Xu, R.~Wang, C.-W. Fu, and J.~Jia, ``Snr-aware low-light image enhancement,'' in {\em CVPR}, 2022.

\bibitem{retinexformer}
Y.~Cai, H.~Bian, J.~Lin, H.~Wang, R.~Timofte, and Y.~Zhang, ``Retinexformer: One-stage retinex-based transformer for low-light image enhancement,'' in {\em ICCV}, 2023.

\bibitem{FourLLIE}
C.~Wang, H.~Wu, and Z.~Jin, ``Fourllie: Boosting low-light image enhancement by fourier frequency information,'' in {\em ACM MM}, 2023.

\bibitem{kind+}
Y.~Zhang, X.~Guo, J.~Ma, W.~Liu, and J.~Zhang, ``Beyond brightening low-light images,'' {\em International Journal of Computer Vision}, 2021.

\bibitem{whjICME}
H.~Wu, H.~Qi, J.~Luo, Y.~Li, and Z.~Jin, ``A lightweight image entropy-based divide-and-conquer network for low-light image enhancement,'' in {\em ICME}, 2022.

\bibitem{brightenandcolorize}
C.~Wang and Z.~Jin, ``Brighten-and-colorize: A decoupled network for customized low-light image enhancement,'' in {\em ACM MM}, 2023.

\bibitem{attentionisall}
A.~Vaswani, N.~Shazeer, N.~Parmar, J.~Uszkoreit, L.~Jones, A.~Gomez, L.~Kaiser, and I.~Polosukhin, ``Attention is all you need,'' {\em NeurIPS}, 2017.

\bibitem{IPT}
H.~Chen, Y.~Wang, T.~Guo, C.~Xu, Y.~Deng, Z.~Liu, S.~Ma, C.~Xu, C.~Xu, and W.~Gao, ``Pre-trained image processing transformer,'' in {\em CVPR}, 2021.

\bibitem{EPT}
W.~Li, X.~Lu, S.~Qian, J.~Lu, X.~Zhang, and J.~Jia, ``On efficient transformer-based image pre-training for low-level vision,'' {\em arXiv preprint arXiv:2112.10175}, 2021.

\bibitem{retinextheory}
E.~H. Land and J.~J. McCann, ``Lightness and retinex theory,'' {\em Josa}, vol.~61, no.~1, pp.~1--11, 1971.

\bibitem{LOLv1}
W.~Y. J.~L. Chen~Wei, Wenjing~Wang, ``Deep retinex decomposition for low-light enhancement,'' in {\em BMVC}, 2018.

\bibitem{mirnet}
S.~W. Zamir, A.~Arora, S.~Khan, M.~Hayat, F.~S. Khan, M.-H. Yang, and L.~Shao, ``Learning enriched features for real image restoration and enhancement,'' in {\em ECCV}, 2020.

\bibitem{he2021srdrl}
Z.~He, Z.~Jin, and Y.~Zhao, ``Srdrl: A blind super-resolution framework with degradation reconstruction loss,'' {\em IEEE TMM}, vol.~24, pp.~2877--2889, 2021.

\bibitem{jin2019flexible}
Z.~Jin, M.~Z. Iqbal, D.~Bobkov, W.~Zou, X.~Li, and E.~Steinbach, ``A flexible deep cnn framework for image restoration,'' {\em IEEE TMM}, vol.~22, no.~4, pp.~1055--1068, 2019.

\bibitem{jin2017robust}
Z.~Jin, T.~Tillo, W.~Zou, Y.~Zhao, and X.~Li, ``Robust plane detection using depth information from a consumer depth camera,'' {\em IEEE TCSVT}, vol.~29, no.~2, pp.~447--460, 2017.

\bibitem{llnet}
K.~G. Lore, A.~Akintayo, and S.~Sarkar, ``Llnet: A deep autoencoder approach to natural low-light image enhancement,'' {\em PR}, vol.~61, pp.~650--662, 2017.

\bibitem{tip18-related}
X.~Liu, G.~Cheung, X.~Ji, D.~Zhao, and W.~Gao, ``Graph-based joint dequantization and contrast enhancement of poorly lit jpeg images,'' {\em IEEE TIP}, vol.~28, no.~3, pp.~1205--1219, 2018.

\bibitem{eccv14-related}
Y.~Li, F.~Guo, R.~T. Tan, and M.~S. Brown, ``A contrast enhancement framework with jpeg artifacts suppression,'' in {\em ECCV}, 2014.

\bibitem{pcm18-related}
C.~Xu, S.~Hao, Y.~Guo, and R.~Hong, ``Enhancing low-light images with jpeg artifact based on image decomposition,'' in {\em PCM}, 2018.

\bibitem{qgac}
M.~Ehrlich, L.~Davis, S.-N. Lim, and A.~Shrivastava, ``Quantization guided jpeg artifact correction,'' in {\em ECCV}, 2020.

\bibitem{fbcnn}
J.~Jiang, K.~Zhang, and R.~Timofte, ``Towards flexible blind jpeg artifacts removal,'' in {\em ICCV}, 2021.

\bibitem{LOLv2}
W.~Yang, W.~Wang, H.~Huang, S.~Wang, and J.~Liu, ``Sparse gradient regularized deep retinex network for robust low-light image enhancement,'' {\em IEEE TIP}, vol.~30, pp.~2072--2086, 2021.

\bibitem{HWMNet}
C.-M. Fan, T.-J. Liu, and K.-H. Liu, ``Half wavelet attention on m-net+ for low-light image enhancement,'' in {\em ICIP}, IEEE, 2022.

\bibitem{restormer}
S.~W. Zamir, A.~Arora, S.~Khan, M.~Hayat, F.~S. Khan, and M.-H. Yang, ``Restormer: Efficient transformer for high-resolution image restoration,'' in {\em CVPR}, 2022.

\bibitem{diffLLIE}
H.~Jiang, A.~Luo, H.~Fan, S.~Han, and S.~Liu, ``Low-light image enhancement with wavelet-based diffusion models,'' {\em ACM TOG}, 2023.

\bibitem{ViT}
A.~Dosovitskiy, L.~Beyer, A.~Kolesnikov, D.~Weissenborn, X.~Zhai, T.~Unterthiner, M.~Dehghani, M.~Minderer, G.~Heigold, S.~Gelly, J.~Uszkoreit, and N.~Houlsby, ``An image is worth 16x16 words: Transformers for image recognition at scale,'' {\em ICLR}, 2021.

\bibitem{charbonnier}
W.-S. Lai, J.-B. Huang, N.~Ahuja, and M.-H. Yang, ``Fast and accurate image super-resolution with deep laplacian pyramid networks,'' {\em IEEE TPAMI}, vol.~41, no.~11, pp.~2599--2613, 2018.

\bibitem{qfrec}
E.~Morley, ``What quality setting should i use for jpg photos?.'' \url{https://www.lenspiration.com/2020/07/what-quality-setting-should-i-use-for-jpg-photos/}, 2020.

\bibitem{ssim}
Z.~Wang, A.~C. Bovik, H.~R. Sheikh, and E.~P. Simoncelli, ``Image quality assessment: from error visibility to structural similarity,'' {\em IEEE TIP}, vol.~13, no.~4, pp.~600--612, 2004.

\bibitem{lpips}
R.~Zhang, P.~Isola, A.~A. Efros, E.~Shechtman, and O.~Wang, ``The unreasonable effectiveness of deep features as a perceptual metric,'' in {\em CVPR}, 2018.

\end{thebibliography}

\end{document}